\documentclass[journal]{IEEEtran}

\ifCLASSINFOpdf
\else
   \usepackage[dvips]{graphicx}
\fi
\usepackage{url}

\usepackage{graphicx}

\newcommand\bftab{\fontseries{b}\selectfont}

\usepackage{graphicx}
\usepackage{amssymb}
\usepackage{booktabs}
\usepackage{comment}
\usepackage{makecell}
\usepackage{float}
\usepackage{multirow}
\usepackage{lipsum}
\usepackage{color, colortbl}
\usepackage{hyperref}
\usepackage[table]{xcolor}
\usepackage{amsmath}
\usepackage{comment}

\begin{document}

\title{Lumos3D: A Single-Forward Framework for Low-Light 3D Scene Restoration}

%\author{First A. Author, \IEEEmembership{Fellow, IEEE}, Second B. Author, and Third C. Author, Jr., \IEEEmembership{Member, IEEE}
\author{Hanzhou Liu, Peng Jiang, Jia Huang and Mi Lu

\thanks{This paragraph of the first footnote will contain the date on which you submitted your paper for review. This research used the DeltaAI advanced computing and data resource, which is supported by the National Science Foundation (award OAC 2320345) and the State of Illinois.}
\thanks{Hanzhou Liu, Peng Jiang, Jia Huang and Mi Lu are with Texas A\&M University, College Station, TX 77801 USA (e-mail: hanzhou1996@tamu.edu; maskjp@tamu.edu; jia.huang@tamu.edu; mlu@ece.tamu.edu).}
}

\markboth{Journal of \LaTeX\ Class Files, Vol. 14, No. 8, August 2015}
{Shell \MakeLowercase{\textit{et al.}}: Bare Demo of IEEEtran.cls for IEEE Journals}
\maketitle

\begin{abstract}
Restoring 3D scenes with low-light conditions is challenging, and most existing methods depend on precomputed camera poses and scene-specific optimization, which greatly restricts their application to real-world scenarios.
To overcome these limitations, we propose Lumos3D, a pose-free single-forward framework for 3D low-light scene restoration.
First, we develop a cross-illumination distillation scheme, where a frozen teacher network takes normal-light ground truth images as input to distill accurate geometric information to the student model.
Second, we define a Lumos loss to improve the restoration quality of the reconstructed 3D Gaussian space.
Trained on a single dataset, Lumos3D performs inference in a purely feed-forward manner, directly restoring illumination and structure from unposed, low-light multi-view images without any per-scene training or optimization.
Experiments on real-world datasets demonstrate that Lumos3D achieves competitive restoration results compared to scene-specific methods.
Our codes are available at https://github.com/HanzhouLiu/Lumos3D.
\end{abstract}


\IEEEpeerreviewmaketitle

\section{Introduction}
\IEEEPARstart{I}{n} recent years, multiple studies~\cite{wang2023lighting,qu2024lush,wang2024bilateral,cui2024aleth,cui2025luminance} have explored adapting Neural Radiance Field (NeRF)~\cite{mildenhall2021nerf} and Gaussian Splatting (3DGS)~\cite{kerbl20233d} to real-world scenarios with challenging illumination~\cite{kwon2025r3evision}.
However, these methods rely on precomputed camera poses and per-scene optimization, which \textbf{makes them difficult to generalize to unseen environments and restore a new 3D scene in a real-time manner}.

In a separate line of research, early neural-network models~\cite{ummenhofer2017demon,zhou2018deeptam,teed2020deepv2d,wang2021multi} have demonstrated the feasibility of end-to-end learning-based 3D reconstruction.
DUSt3R~\cite{wang2024dust3r} and MASt3R~\cite{leroy2024grounding} directly predict geometry from a pair of unposed views.  
Later, Spann3R~\cite{wang20253d}, CUT3R~\cite{wang2025continuous}, and MUSt3R~\cite{cabon2025must3r} further reduce reliance on classical optimization. 
More recently, VGGT~\cite{wang2025vggt} introduces a novel Transformer architecture for joint multi-view inference of depth, pose, and point maps.
AnySplat~\cite{jiang2025anysplat} extends VGGT into an efficient and real-time feed-forward 3DGS framework.
However, \textbf{the extension of single-forward 3D reconstruction methods to low-light restoration has not yet been explored}.

To this end, we propose \textbf{\textit{Lumos3D}}, a pose-free single-forward framework that restores illumination and structure from unposed multi-view low-light inputs.
Lumos3D is trained once on a dataset with synthetic degradations and performs low-light 3D scene restoration in a single forward pass, while achieving competitive restoration quality against scene-specific methods such as Aleth-NeRF~\cite{cui2024aleth} and Lumincance-GS~\cite{cui2025luminance}.

With VGGT~\cite{wang2025vggt} as the geometry backbone, Lumos3D first estimates geometric cues including depth and camera poses from low-light inputs. 
Different from the prior VGGT-based distillation strategies~\cite{jiang2025anysplat,liu2025stylos}, 
our \textbf{\textit{cross-illumination distillation scheme}} uses a frozen teacher network operating on the normal-light ground-truth to provide geometric supervision,
while the trainable student model learns from low-light context images. 
This novel strategy offers cleaner and more reliable geometric guidance by leveraging paired normal-light and low-light observations during the training.

%Then, Lumos3D reconstructs a 3D Gaussian scene from the estimated geometry, following the explicit representation pipeline of AnySplat~\cite{jiang2025anysplat}. 
%A dual-head decoder predicts per-pixel depth and confidence maps, which are back-projected into 3D space using the estimated camera poses to obtain Gaussian centers. 
%Meanwhile, a Gaussian head regresses opacity, orientation, scale, and spherical harmonic color coefficients, producing an explicit and differentiable 3D Gaussian representation that can be rendered through Gaussian rasterization~\cite{kerbl20233d}. 
%This explicit formulation enables efficient end-to-end optimization while preserving geometric fidelity.

To further improve the restoration quality, 
we design a \textbf{\textit{Lumos loss}}, 
composed of a content loss, an image-level $\ell_1$ loss, and a voxel-level statistical loss. 
Combined with the proposed cross-illumination distillation, 
Lumos3D enables geometry-aware illumination restoration and produces high-quality rendered results that are competitive with recent scene-specific models Aleth-NeRF~\cite{cui2024aleth} and Lumincance-GS~\cite{cui2025luminance}.

%Our main contributions are three-fold:
%\begin{itemize}
%    \item We propose \textit{Lumos3D}, a single-forward framework for low-light 3D scene restoration, without any precomputed camera pose or per-scene optimization required.
%    \item We introduce a \textit{cross-illumination distillation} scheme that transfers geometric knowledge from normal-light supervision to low-light learning, enabling more reliable depth reasoning under challenging illumination.
%    \item We propose \textit{Lumos losses} that jointly enforce photometric and geometric consistency for robust and coherent low-light 3D scene restoration.
%\end{itemize}
%Collectively, these contributions establish a new paradigm for low-light 3D scene restoration, enabling pose-free reconstruction without any per-scene training in a single forward pass.

%\input{section02_review}

\section{Methodology}
\begin{figure*}[tb]
  \centering
  % left, bottom, right, top
  \includegraphics[width=0.9\linewidth, trim = 7.5cm 5.5cm 6.4cm 5.5cm]{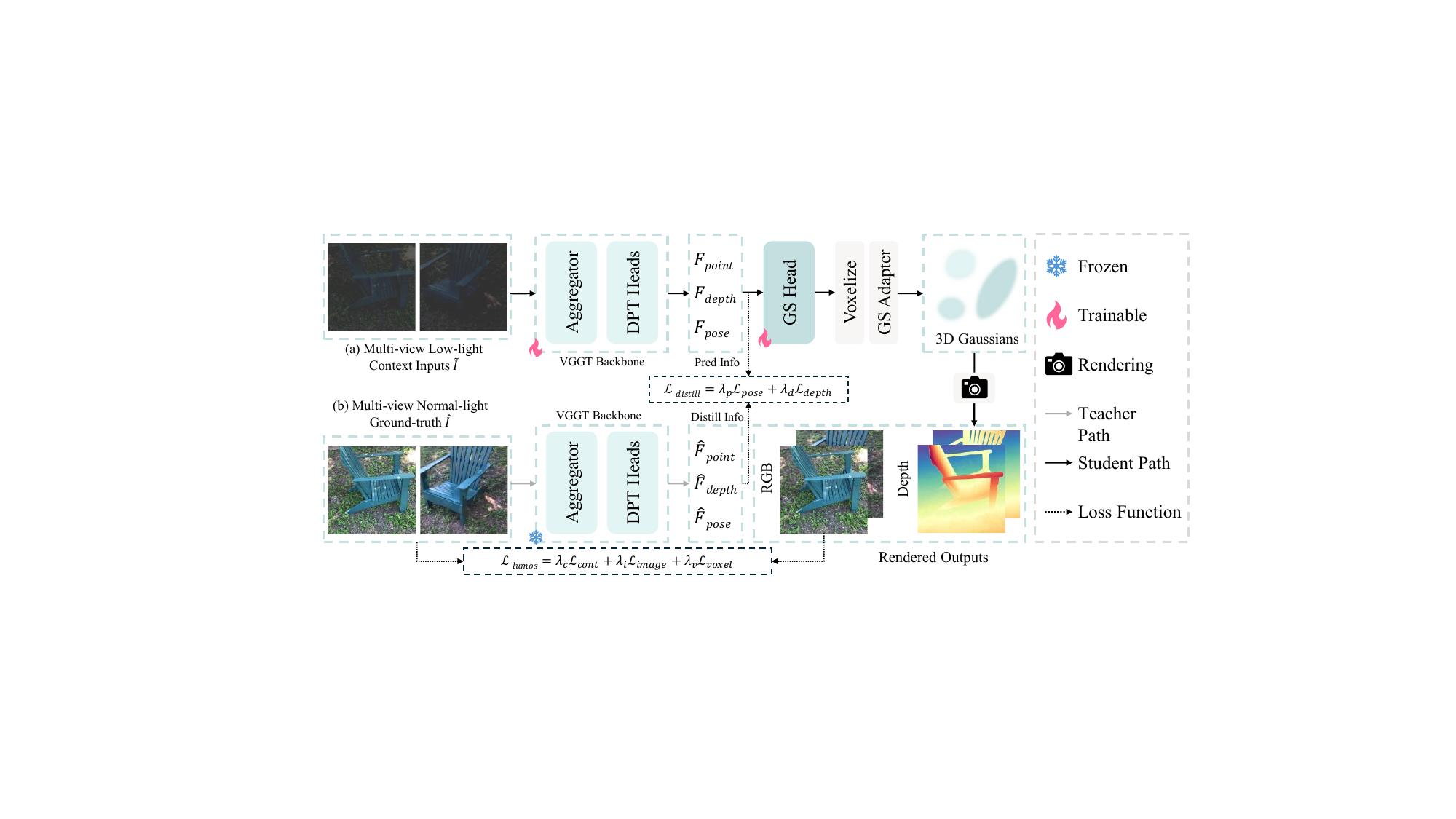}
  %\fbox{\includegraphics[width=\linewidth, trim = 7.5cm 5.5cm 6.4cm 5.5cm]{arch.pdf}}
  \caption{Architecture overview. Given multi-view low-light images, \textbf{Lumos3D} predicts the restored 3D Gaussian representation in a single-forward way.}\label{fig:arch}
\end{figure*}

\subsection{Problem Formulation}
Given low-light multi-view images $\tilde{I}$, the network predicts, in a single forward pass, the restored 3D Gaussian representation $\mathcal{G}$.
Formally, the process is expressed as,
\begin{equation}
\mathcal{G} = 
\Phi_\theta\big(\tilde{I}\big),
\end{equation}
where $\Phi_\theta$ denotes the proposed network parameterized by $\theta$.

\subsection{Pipeline}
\noindent\textbf{Architecture Overview.}  \
As shown in Fig.~\ref{fig:arch}, 
Lumos3D first extracts geometry-aware features from low-light inputs and estimates per-view poses, depth maps, and point maps, which are then transformed into a set of pixel-wise Gaussian primitives of the current scene.  
After that, Lumos3D applies differentiable voxelization and produces voxel-wise Gaussian primitives~\cite{jiang2025anysplat}, which are used to reconstruct the restored 3D Gaussian scene and render high-quality RGB images.

\medskip
\noindent\textbf{Training Objective.}
During training, paired normal-light and synthetic low-light multi-view images $\{\hat{I}, \tilde{I}\}$ are provided.
A frozen teacher network operating on normal-light inputs provides geometric supervision for the student model trained on low-light images.
Lumos3D is optimized with a unified objective consisting of three components: 
(i) a reconstruction loss enforcing fidelity between restored images and normal-light ground truth, 
(ii) a distillation loss transferring geometric priors from the teacher to the student, and 
(iii) the proposed Lumos loss, which regularizes illumination-aware structural stability in the reconstructed 3D representation,
\begin{equation}
L_{\text{total}} =
L_{\text{rec}} +
\omega_{\text{distill}} \, \mathcal{L}_{\text{distill}} +
\omega_{\text{lumos}} \,  \mathcal{L}_{\text{lumos}}.
\end{equation}
For brevity, we omit averaging over batch, spatial, and channel dimensions in this manuscript due to space limits.

\subsection{Cross-Illumination Distillation}\label{sec:distill_loss}
As shown in Fig.~\ref{fig:arch}, we leverage a teacher–student framework, in which a teacher network frozen under normal illumination, serves as a stable source of geometry-rich and illumination-invariant supervision. By contrast, the student network operates directly on low-light inputs and learns to approximate the teacher’s predictions despite the degradation in visibility. This setup allows the student to inherit structural reasoning capabilities from the teacher while simultaneously adapting to the challenges posed by low-light conditions. Specifically, we distill both camera poses and depth information using the following loss function,
\begin{equation}
\mathcal{L}_{\text{distill}} =
\frac{1}{v}
\sum_{v=1}^{V}
\left\|
\hat{F}_{pose}^{(v)} - 
F_{pose}^{(v)}
\right\|_h
+
\frac{1}{v}
\sum_{v=1}^{V}
(
\hat{D}^{(v)} - 
D^{(v)}
)^2.
\end{equation}
Here, $V$ is the number of views; each $\hat{F}_{pose}^{(v)}$ represents the pseudo ground-truth pose encoding obtained from the pre-trained VGGT~\cite{wang2025vggt} on normal-light context images, while $F_{pose}^{(v)}$ is the pose encoding estimated by the student model on corresponding low-light context inputs; $\hat{D}^{(v)}$ denotes the pseudo depth map, while $D^{(v)}$ is the depth map rendered by the student model; $\|\cdot\|_{h}$ denotes the huber loss.

\subsection{Lumos Loss}\label{sec:lumos_loss}
To further improve the restoration quality under low-light conditions, 
we define $\mathcal{L}_{\text{Lumos}}$ as,
\begin{equation}
\mathcal{L}_{\text{Lumos}}
=
\lambda_c \, \mathcal{L}_{\text{content}}
+
\lambda_i \, \mathcal{L}_{\text{image}}
+
\lambda_v \, \mathcal{L}_{\text{voxel}},
\end{equation}
where $\lambda_c$, $\lambda_i$, and $\lambda_v$ are the weighting coefficients for the content-, image-, 
and voxel-level losses, respectively, with default settings of $0.1$, $1.0$, and $0.01$.  

\subsubsection{Content-Level Feature Loss}
The content loss encourages high-level semantic consistency between the rendered multi-view images $I$ and the normal-light ground-truth $\hat{I}$. 
Specifically, we extract intermediate features from a pretrained VGG~\cite{simonyan2015very} network and compute their $\ell_1$ difference following,
\begin{align}
\mathcal{L}_{\text{content}} 
&= 
\frac{1}{V}
\sum_{v=1}^{V}
\left\| 
\hat{F}^{(v)}_{VGG} - F^{(v)}_{VGG}
\right\|_1,
\end{align}
where $\hat{F}^{(v)}_{VGG}$ and $F^{(v)}_{VGG}$ represent the VGG feature vectors 
of the ground-truth and rendered images respectively; $\|\cdot\|_1$ measures the element-wise difference between the two features.

\subsubsection{Image-Level Restoration Loss}
To guarantee pixel-wise accuracy, we adopt an $\ell_1$ loss between the rendered multi-view images $I$ 
and the normal-light ground-truth $\hat{I}$,
\begin{align}
\mathcal{L}_{\text{image}} 
&= 
\frac{1}{V}
\sum_{v=1}^{V}
\left\| 
\hat{I}^{(v)} - I^{(v)}
\right\|_1.
\end{align}

\subsubsection{Voxel-Level 3D Consistency Loss}
To enforce geometric coherence across multi-view observations, we introduce a voxel-level 3D consistency loss.
Following the voxelization paradigm introduced in AnySplat~\cite{jiang2025anysplat} and Stylos~\cite{liu2025stylos}, multi-view 2D features extracted at multiple scales from the rendered images and the normal-light ground-truth images are back-projected and fused into a shared 3D voxel grid using the estimated geometry.
Then, we align the mean and variance statistics between these voxelized features across scales,
\begin{align}
\mathcal{L}_{\text{voxel}} 
&= 
\sum_{i=1}^{5} 
w_i
\left(
\left\| 
\hat{\mu}_i - \mu_i 
\right\|_1 
+
\left\| 
\hat{\sigma}_i - \sigma_i 
\right\|_1
\right),
\end{align}
where $\hat{\mu}_i$ and $\hat{\sigma}_i$ denote the mean and standard deviation of voxelized features from the teacher branch at scale $i$, and $\mu_i$, $\sigma_i$ correspond to those from the student branch.  
The scale weights $w_i$ are normalized such that $\sum_{i=1}^{5} w_i = 1$.  

\begin{figure*}[th]
    \centering
    % left, bottom, right, top
    \includegraphics[width=0.98\linewidth, trim = 0.22cm 11.1cm 12.5cm 0.22cm]{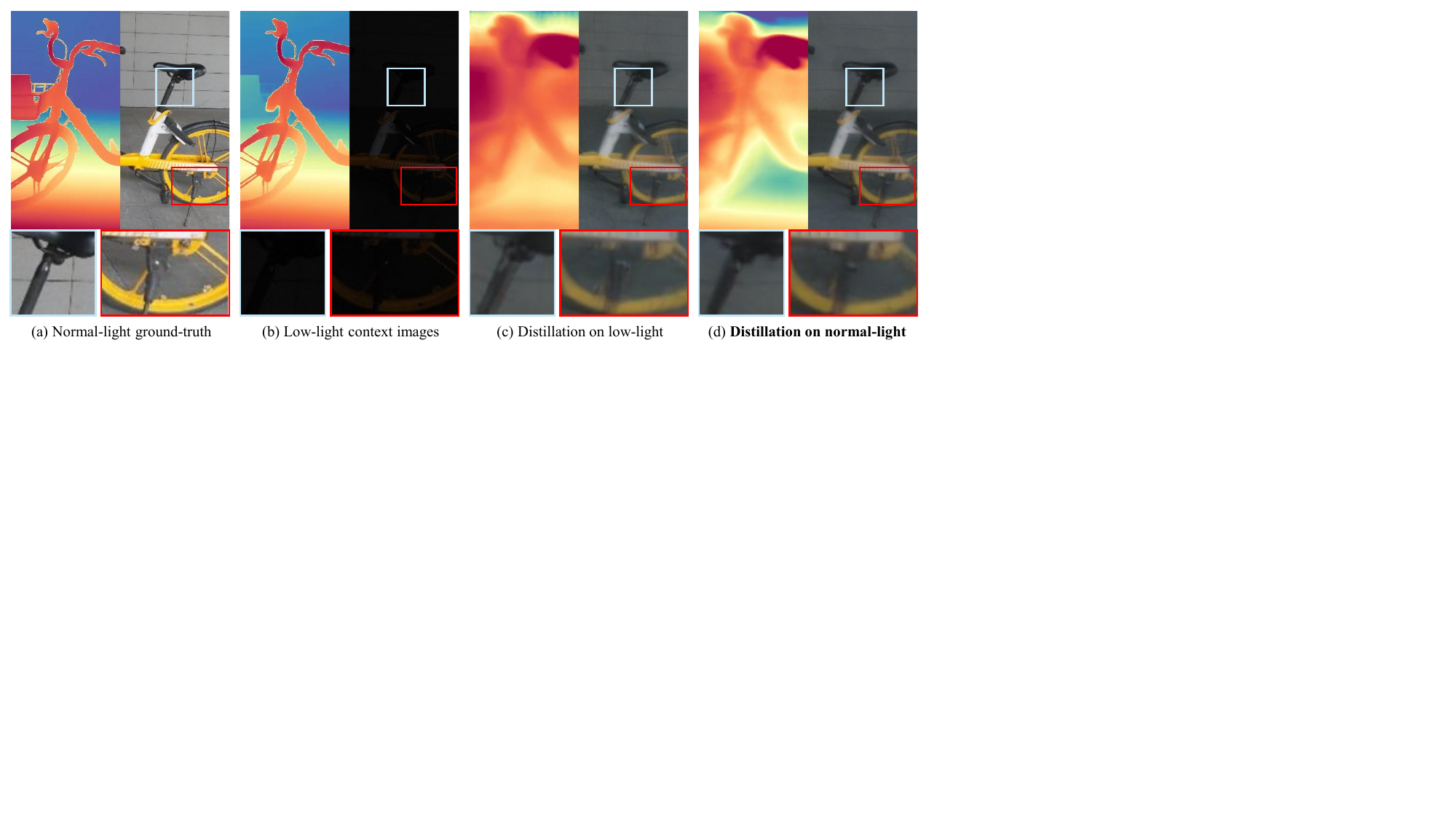}
    %\fbox{\includegraphics[width=0.98\linewidth, trim = 0.22cm 11.1cm 12.5cm 0.22cm]{distillation.pdf}}
    \caption{
    Qualitative comparison of different distillation schemes. 
    Each visualization corresponds to the same scene, with depth on the left and the corresponding RGB image on the right. 
    In the depth maps, blue denotes distant regions and red denotes nearby ones. 
    Distillation on low-light images suffers from illumination ambiguity, 
    whereas distillation on normal-light images yields more accurate and geometrically cleaner depth and relighting results.
    }
    \label{fig:qualitative_distill}
\end{figure*}

\section{Experiment}
\begin{table}[tb]
  \setlength{\tabcolsep}{4.6pt}
  \centering
  \caption{
    Ablation on distillation targets of the teacher model and Lumos loss variants.
    Quantitative results are averaged across the \textbf{LOM} dataset using models trained on \textbf{DL3DV}.
  }
  \begin{tabular}{c c c c c c c c}
    \toprule
    \multicolumn{2}{c}{Distillation} &
    \multicolumn{3}{c}{Lumos losses} &
    \multicolumn{3}{c}{Metrics (Average)} \\
    \cmidrule(lr){1-2}\cmidrule(lr){3-5}\cmidrule(lr){6-8}
    Low$^1$ & GT$^{2}$ &
    Content & Image & Voxel &
    PSNR$\uparrow$ & SSIM$\uparrow$ & LPIPS$\downarrow$ \\
    \midrule
    \checkmark &  &  &  &  & \bftab17.93 & 0.758 & 0.405 \\
    \rowcolor{gray!30}
     & \checkmark &  &  &  & 17.47 & \bftab0.758 & \bftab0.402 \\
    \midrule
     & \checkmark &  &  &  & 17.47 & 0.758 & 0.402 \\
     & \checkmark & \checkmark &  &  & 17.59 & 0.765 & 0.403 \\
     & \checkmark & \checkmark & \checkmark &  & 19.41 & 0.782 & 0.402 \\
     \rowcolor{gray!30}
     & \checkmark & \checkmark & \checkmark & \checkmark & \bftab19.76 & \bftab0.784 & \bftab0.396 \\
    \bottomrule
  \end{tabular}
  
  \vspace{3pt}
  %\footnotesize
  \textit{Note.} 
  $^{1}$ Low = low-light context images; 
  $^{2}$ GT = normal-light ground-truth.
  The symbol $\uparrow$ indicates that a higher score reflects better performance for the corresponding metric, whereas $\downarrow$ indicates that a lower score represents better performance. $\checkmark$ represents the method used in that experiment.
  \label{tab:ablation_study_img_quality}
\end{table}
\begin{table*}[t]
\centering
\caption{Quantitative comparison of different models on the \textbf{LOM} dataset. Best results are highlighted in \bftab{bold}.}
\begin{tabular}{l|ccc|ccc|ccc|ccc}
\toprule
\multirow{2}{*}{Models} 
& \multicolumn{3}{c|}{Bike} 
& \multicolumn{3}{c|}{Buu} 
& \multicolumn{3}{c|}{Chair} 
& \multicolumn{3}{c}{Sofa} \\ 
\cmidrule(lr){2-4} \cmidrule(lr){5-7} \cmidrule(lr){8-10} \cmidrule(lr){11-13}
& PSNR$\uparrow$ & SSIM$\uparrow$ & LPIPS$\downarrow$
& PSNR$\uparrow$ & SSIM$\uparrow$ & LPIPS$\downarrow$
& PSNR$\uparrow$ & SSIM$\uparrow$ & LPIPS$\downarrow$ 
& PSNR$\uparrow$ & SSIM$\uparrow$ & LPIPS$\downarrow$ \\ 
\midrule
\multicolumn{13}{l}{\textbf{Low-light}} \\ 
\midrule
Aleth\_NeRF~\cite{cui2024aleth}    
& \bftab 16.50 & \bftab 0.661 & 0.481 & 16.52 & 0.707 & \bftab 0.418 & 16.54 & 0.768 & \bftab 0.536 & 16.53 & 0.805 & 0.408 \\
Luminance-GS~\cite{cui2025luminance}
& 16.39 & 0.627 & 0.520 & 15.40 & 0.725 & 0.436 & \bftab 18.58 & 0.690 & 0.634 & 18.98 & 0.756 & 0.472 \\
\rowcolor{gray!20}
Lumos3D (Ours)    
& 14.07 & 0.605 & \bftab 0.432 & \bftab 19.16 & \bftab 0.755 & 0.420 & 17.82 & \bftab 0.781 & 0.565 & \bftab 22.21 & \bftab 0.848 & \bftab 0.346 \\
\midrule
\multicolumn{13}{l}{\textbf{Over-exposure}} \\ 
\midrule
Aleth\_NeRF~\cite{cui2024aleth}    
& 19.02 & 0.705 & 0.423 & 15.16 & 0.709 & 0.682 & 19.02 & 0.789 & 0.545 & 18.14 & 0.822 & 0.459 \\
Luminance-GS~\cite{cui2025luminance}
& 19.72 & 0.646 & 0.365 & \bftab 15.66 & \bftab 0.729 & 0.511 & 20.16 & 0.670 & 0.392 & 19.59 & 0.751 & 0.410 \\
\rowcolor{gray!20}
Lumos3D (Ours)      
& \bftab 20.92 & \bftab 0.733 & \bftab 0.289 & 15.00 & 0.711 & \bftab 0.493 & \bftab 21.99 & \bftab 0.790 & \bftab 0.386 & \bftab 22.37 & \bftab 0.847 & \bftab 0.339 \\
\bottomrule
\end{tabular}
\label{tab:lowlight_overexposure}
\end{table*}
\begin{figure*}[tb]
    \centering
    % left, bottom, right, top
    \includegraphics[width=0.98\linewidth, trim = 0.22cm 3.1cm 8.9cm 0.22cm]{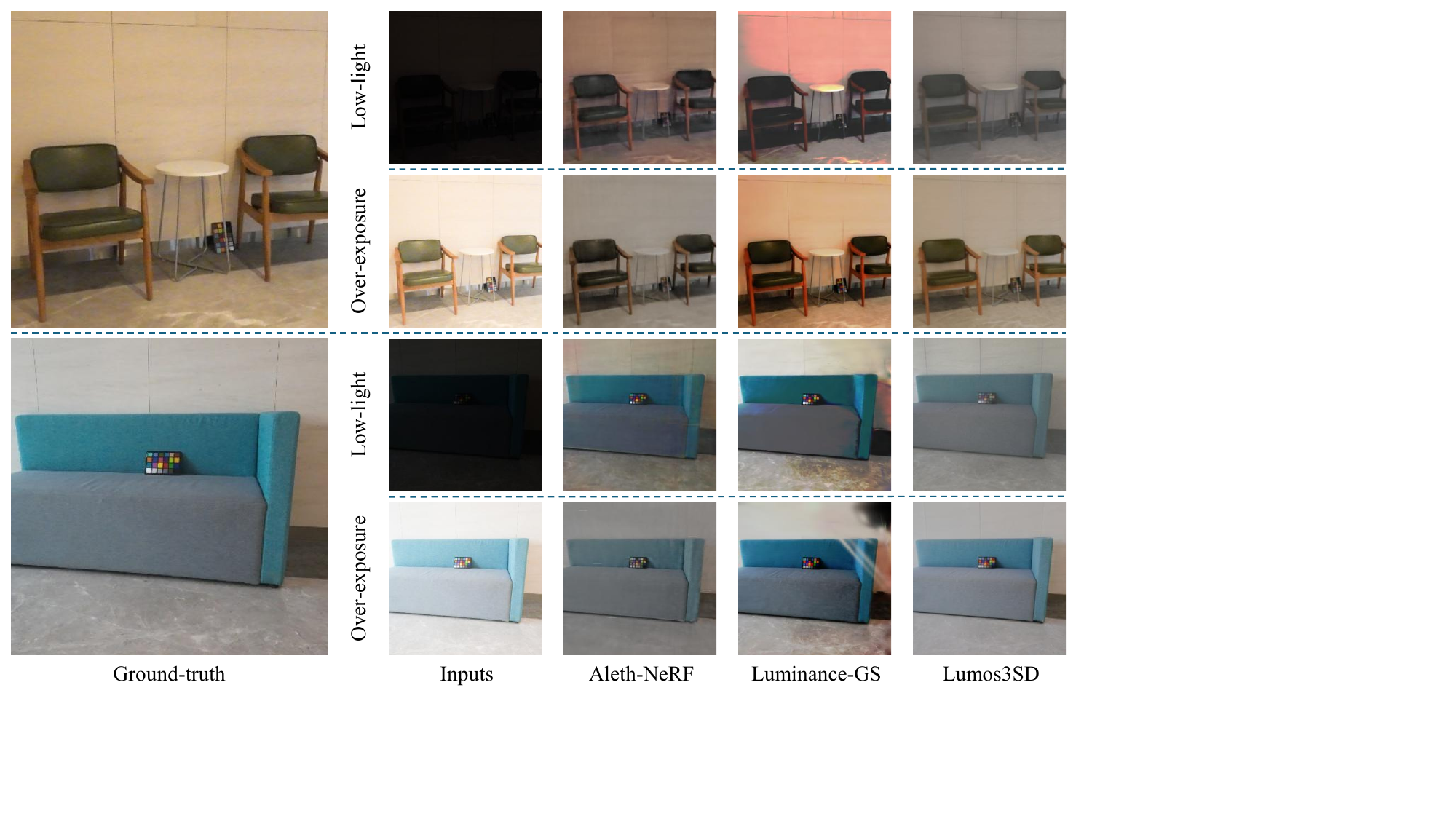}
    %\fbox{\includegraphics[width=0.98\linewidth, trim = 0.22cm 3.1cm 8.9cm 0.22cm]{sota.pdf}}
    \caption{
    Qualitative comparison of different 3D low-light and over-exposure restoration schemes on the chair and sofa scenes in the LOM dataset.
    }
    \label{fig:sota}
\end{figure*}
\noindent\textbf{Datasets.}
We use DL3DV~\cite{ling2024dl3dv} as the training set. 
To simulate low-light inputs, we randomly scale the exposure by a factor between 0.05 and 0.1, and apply a gamma correction of 1.3–1.4 in the linear RGB domain. 
For evaluation, we use the bike, buu, chair and sofa scenes in the LOM dataset~\cite{cui2024aleth}.

\medskip
\noindent\textbf{Implementation Details.} 
We train our network with a dynamic batch size of 22, corresponding to the maximum number of views per GPU. 
The entire training process consists of 30K iterations. 
The initial learning rate is set to $2\times10^{-4}$ and follows a cosine annealing schedule with a warm-up phase of 1K iterations. 
Training converges within approximately 60 hours using eight NVIDIA GH200 GPUs on two nodes.

\medskip
\noindent\textbf{Evaluation Metrics.}
Following previous related works~\cite{cui2024aleth,cui2025luminance}, we report PSNR, SSIM, and LPIPS~\cite{zhang2018unreasonable} between the predicted rendered images and normal-light ground-truth. 

\subsection{Ablation Study}

For ablation studies, all the model variants are trained on the first 6K scenes of the DL3DV dataset with a dynamic batch size of 18 for up to 20,000 steps using four NVIDIA GH200 GPUs. Throughout this section, the terms \textit{normal-light} and \textit{ground truth} are used interchangeably, whereas \textit{low-light} and \textit{context images} denote the same input modality.

\subsubsection{Distillation}
While Table~\ref{tab:ablation_study_img_quality} shows that distillation using normal-light ground truth results in lower PSNR compared to low-light–based distillation, qualitative results in Fig.~\ref{fig:qualitative_distill} reveal clear advantages in geometric accuracy.
Depth visualizations for both normal-light and low-light inputs are generated using Depth Anything V2~\cite{yang2024depth}.
The model distilled with normal-light supervision produces more accurate depth estimates and cleaner 3D reconstructions.
This observation is consistent with the intuition that normal-light inputs provide more reliable geometric cues.
Moreover, SSIM remains comparable while LPIPS shows a slight improvement.
Therefore, we adopt the ground-truth–based distillation as the default configuration, prioritizing geometric fidelity over marginal PSNR gains.

%For instance, Figure~\ref{fig:qualitative_distill} shows that the supporting rod beneath the bicycle saddle exhibits no ghosting, and the wheel contours appear smoother. Therefore, we adopt the ground-truth–based distillation as the default configuration. 

\subsubsection{Losses}
As shown in Table~\ref{tab:ablation_study_img_quality}, using the baseline reconstruction loss produces PSNR of 17.47 dB. Adding the content-level loss offers a small but noticeable improvement, raising PSNR to 17.59. Incorporating the image-level loss leads to a more significant boost: PSNR jumps to 19.41. Finally, introducing the voxel-level loss yields the best overall results, achieving 19.76 dB PSNR. Taken together, these ablations show that each Lumos component contributes incrementally, while their full combination delivers the strongest reconstruction quality under low-light conditions.
%As shown in Table~\ref{tab:ablation_study_img_quality}, using only the baseline reconstruction loss produces 17.47 dB PSNR, 0.758 SSIM, and 0.402 LPIPS, serving as the reference point for subsequent variants. Adding the content-level loss offers a small but consistent improvement, raising PSNR to 17.59 and SSIM to 0.765, though LPIPS slightly increases to 0.403, suggesting that semantic alignment alone mainly benefits structural consistency. Incorporating the image-level loss leads to a more significant boost: PSNR jumps to 19.41 and SSIM to 0.782, while LPIPS remains stable at 0.402, confirming the advantage of pixel-space illumination guidance in enhancing global brightness and contrast. Finally, introducing the voxel-level loss yields the best overall results, achieving 19.76 dB PSNR, 0.784 SSIM, and the lowest LPIPS of 0.396, indicating improved multi-view geometric coherence and perceptual fidelity. Taken together, these ablations show that each Lumos component contributes incrementally, while their full combination delivers the strongest reconstruction quality under low-light conditions.

\subsection{Comparison with State-of-the-Art Methods}

To the best of our knowledge, no existing approach offers a single-forward solution for low-light 3D scene restoration. Prior methods typically rely on per-scene optimization. 
\
We therefore compare our Lumos3D with scene-specific methods \textit{Aleth-NeRF}~\cite{cui2024aleth} and \textit{Luminance-GS}~\cite{cui2025luminance}. 
As shown in Table~\ref{tab:lowlight_overexposure}, Lumos3D achieves highly competitive performance without any scene-specific adaptation. 
Furthermore, we demonstrate that Lumos3D can be readily extended to address other illumination degradations, such as over-exposure restoration.

\medskip
\noindent \textbf{Over-exposure.} 
As shown in Table~\ref{tab:lowlight_overexposure}, despite being trained exclusively on synthetic over-exposure image pairs, Lumos3D generalizes effectively to real-world high-exposure scenes without any fine-tuning, achieving compelling performance.
%For example, on the \textit{Bike} scene, \textbf{Lumos3D} attains \textbf{20.92\,dB} PSNR and \textbf{0.733} SSIM, surpassing \textit{Aleth-NeRF} by +1.9\,dB and exhibiting significantly better perceptual quality, LPIPS $\downarrow$ from 0.423 to \textbf{0.289}. 
%Despite being trained exclusively on synthetic over-exposure data, Lumos3D generalizes effectively to real-world high-exposure scenes without any fine-tuning.

\section{Conclusion}
In this work, we presented \textbf{Lumos3D}, a single-forward framework for low-light 3D scene restoration. 
Unlike prior scene-specific approaches that rely on precomputed camera poses and per-scene optimization, Lumos3D reconstructs and restore 3D scenes from unposed multi-view low-light inputs without any per-scene fitting required. 
The proposed cross-illumination distillation scheme and Lumos loss significantly improves the 3D scene restoration quality. 
Extensive experiments demonstrate that Lumos3D achieves geometrically accurate AND visually pleasant results, even when trained solely on synthetic data. 
Beyond low-light scenarios, the framework also generalizes to other challenging illumination conditions, such as over-exposure. 
The proposed Lumos3D establishes a new foundation for scalable, optimization-free 3D scene restoration, paving the way toward unified and real-time relighting systems.

\newpage
\bibliography{ref}
\bibliographystyle{IEEEtran}

\end{document}